\documentclass[twoside,11pt]{article}

\usepackage{jmlr2e}
\usepackage[utf8]{inputenc} 
\usepackage[T1]{fontenc}    
\usepackage{hyperref}       
\usepackage{url}            
\usepackage{booktabs}       
\usepackage{amsfonts}       
\usepackage{nicefrac}       
\usepackage{microtype}      
\usepackage[table]{xcolor}

\usepackage{times}
\usepackage{epsfig}
\usepackage{graphicx}
\usepackage{amssymb}
\usepackage{amsmath}

\newcommand{\firstbest}[1]{\textcolor{red}{#1}}
\newcommand{\secondbest}[1]{\textcolor{blue}{#1}}

\newcommand{\spo}[3]{{\small{<$\mathtt{#1, {\color{gray}#2}, #3}$>}}}

\jmlrheading{1}{2000}{1-48}{4/00}{10/00}{}

\ShortHeadings{CREPE: Learnable Prompting With CLIP Improves Visual Relationship Prediction}{CREPE: Learnable Prompting With CLIP Improves Visual Relationship Prediction}
\firstpageno{1}

\begin{document}

\title{CREPE: Learnable Prompting With CLIP Improves Visual Relationship Prediction}

\author{\name Rakshith Subramanyam \email rsubra17@asu.edu \\
       \addr Arizona State University\\
       Tempe, AZ 85281, USA
       \AND
       \name T. S. Jayram \email thathachar1@llnl.gov \\
       \addr Center for Applied Scientific Computing\\
       Lawrence Livermore National Laboratory
       Livermore, CA 94550, USA
       \AND
       \name Rushil Anirudh \email anirudh1@llnl.gov \\
       \addr Center for Applied Scientific Computing\\
       Lawrence Livermore National Laboratory
       Livermore, CA 94550, USA
       \AND
       \name Jayaraman J. Thiagarajan \email jjayaram@llnl.gov \\
       \addr Center for Applied Scientific Computing\\
       Lawrence Livermore National Laboratory
       Livermore, CA 94550, USA}

\editor{}

\maketitle

\begin{abstract}
In this paper, we explore the potential of Vision-Language Models (VLMs), specifically CLIP, in predicting visual object relationships, which involves interpreting visual features from images into language-based relations. Current state-of-the-art methods use complex graphical models that utilize language cues and visual features to address this challenge. We hypothesize that the strong language priors in CLIP embeddings can simplify these graphical models paving for a simpler approach. We adopt the UVTransE relation prediction framework, which learns the relation as a translational embedding with subject, object, and union box embeddings from a scene. We systematically explore the design of CLIP-based subject, object, and union-box representations within the UVTransE framework and propose CREPE (CLIP Representation Enhanced Predicate Estimation). CREPE utilizes text-based representations for all three bounding boxes and introduces a novel contrastive training strategy to automatically infer the text prompt for union-box. Our approach achieves state-of-the-art performance in predicate estimation, mR@5 27.79, and mR@20 31.95 on the Visual Genome benchmark, achieving a $15.3\%$ gain in performance over recent state-of-the-art at mR@20. This work demonstrates CLIP's effectiveness in object relation prediction and encourages further research on VLMs in this challenging domain.

\end{abstract}

\section{Introduction}
Large-scale pretraining has emerged as a dominant technique~\citep{chen2020simple, caron2021emerging} in numerous vision and language processing applications. The advent of multimodal joint embeddings, exemplified by models like CLIP~\citep{radford2021learning}, has further elevated the performance of reasoning tasks involving both vision and language. This paper delves into the practicality of employing vision-language models (VLM) to tackle the challenging problem of automatically detecting the relationship between two objects within a given scene. This capability plays a crucial role in empowering AI systems with situational awareness during tasks such as navigation and environment monitoring.

Despite the advances in object detection, this problem continues to be challenging due to the inherent difficulty in mapping visual features to a broad category to human-understandable predicates; examples range from describing relative spatial locations, indicating action, revealing semantic relations or even providing comparisons. As one would expect, the most successful solutions construct graphical models (e.g., VCTree~\citep{tang2019learning}, GB-Net~\citep{zareian2020bridging}) that leverage both language priors and visual features to create a concise representation of a scene, i.e., objects and their interactions. Furthermore, the long-tail nature of predicate occurrence in even well-curated benchmarks often necessitates the need to explicitly handle severe class imbalance. This practical challenge has led to the emergence of a variety of model architectures, training strategies, and calibration techniques.

Amidst all recent advances in object relation (or predicate) estimation~\citep{zaidi2022survey}, a noteworthy observation is that VLMs such as CLIP \emph{do not} feature in any of the state-of-the-art solutions to the problem. This is counter to the current trend of rapid adoption in a host of multimodal applications. Further, given the fact that CLIP embeddings contain strong language priors by design, we hypothesize that they should be able to simplify graphical models used in existing approaches without sacrificing on performance. In this study, we use UVTransE~\citep{hung2020contextual} combined with CLIP embeddings to verify this hypothesis. UVTransE is one of the earliest predicate estimation methods that does not use sophisticated priors to combine vision and language priors. In a nutshell, it
learns the predicate as a translational embedding, i.e., $\mathrm{p} = \mathrm{u} - (\mathrm{s} + \mathrm{o})$, where $\mathrm{s}$, $\mathrm{o}$ and $\mathrm{u}$ represent the subject, object, and union box embeddings respectively, from a scene.

\emph{But somewhat surprisingly, we find that merely replacing existing object representations (e.g. Faster-RCNN \citep{ren2015faster}) with CLIP embeddings does not provide any boost in performance.}

Motivated by the above, in this paper we systematically study the design of CLIP-based subject, object, and union-box representations within the conventional UVTransE framework. We propose \textbf{CREPE} (\textbf{C}LIP \textbf{R}epresentation \textbf{E}nhanced \textbf{P}redicate \textbf{E}stimation), that produces state-of-the-art performance in predicate estimation without requiring any advanced model architectures or customized training strategies. CREPE advocates for the use of text-based representations for all three bounding boxes, and in particular, proposes to automatically infer the text prompt for each union-box using a novel contrastive training strategy. In order to achieve sufficient visual grounding during prompt generation, we devise a negative sampling strategy based on cross-modal retrieval in the CLIP embedding space. Finally, to tackle the long-tail challenge, we employ a simple prediction correction step that does not involve any learnable parameters or modifications to the optimization process.

Using experiments on the challenging, and widely adopted visual genome (VG) benchmark, we show that CREPE leads to state-of-the-art performance -- achieving \textbf{mR@5$=27.79$} and \textbf{mR@20$=31.95$}. This corresponds to a $15.3\%$ relative gain in mR@20 performance over recent state-of-the-art (VCTree + SCR~\citep{kang2023skew}). Most strikingly, the mR@5 performance of CREPE is even marginally better than the best known mR@20 result in the literature. CREPE also produces a consistently strong performance at both the head and tail ends of the predicate occurrence distribution, validating our hypothesis. 

Overall, our work provides the first evidence for the effectiveness of CLIP in object relation prediction and motivates further research on the utility of VLMs in this challenging domain.

\section{Background}
\label{background}
We consider the problem of visual object relationship prediction, the goal of which is to estimate the relationship between a pair of objects in an image. Such relationships, or predicates, can be spatial, action-based, semantic, or comparative in nature and are commonly represented as triplets in the form of \spo{subject}{predicate}{object}, for example \spo{person}{riding}{bike}. Typically, identifying the predicate first requires an object detection step (for e.g. using Faster R-CNN~\citep{ren2015faster}), followed by a predicate estimation step. Most recent approaches have innovated on better ways to estimate the predicate given the object labels and the corresponding image-level regions \citep{zhu2022scene}. These methods are closely related to (and build on top of) techniques that address knowledge graph embedding (KGE) for relational data, where there is a similar goal of identifying the relationship in facts that are represented as \spo{head}{label}{tail} \citep{bordes2013translating, he2015learning, sun2019rotate}. 

Different from KGE approaches, in the visual domain, we also have visual information about the objects and their relational context that can be taken into account. Generally speaking, there are two key objectives in setting up an effective solution -- (a) the choice of embedding space in which to define the relationship, and (b) the prior relationship between subject/object embeddings and the predicate embedding. If we operate entirely in the visual domain,  the embeddings can be the feature representations for the subject, object, and union box obtained from the object detector, which is expected to contain the predicate information. A popular prior for the relationship is the translational model (inspired from TransE \citep{bordes2013translating}). For subject, object and union box features given as $\mathtt{\mathrm{s}_{\footnotesize{\text{img}}}, \mathrm{o}_{\footnotesize{\text{img}}}, \mathrm{u}_{\footnotesize{\text{img}}}}\in \mathbb{R}^n$, mapped onto a $d$ embedding space using three learnable nonlinear projection functions $f_s$, $f_o$, and $f_u$ -- the likelihood of predicate class is given by: 
\begin{equation}
P(\mathtt{p|\{\mathrm{u}_{\footnotesize{\text{img}}},\mathrm{s}_{\footnotesize{\text{img}}}, \mathrm{o}_{\footnotesize{\text{img}}}\}}) = f_p(f_u(\mathtt{\mathrm{u}_{\footnotesize{\text{img}}}}) - f_s(\mathtt{\mathrm{s}_{\footnotesize{\text{img}}}}) -f_o(\mathtt{\mathrm{o}_{\footnotesize{\text{img}}}})),
\end{equation}
where $f_p$ is a predicate classifier with a softmax activation, that estimates the likelihood of the predicate embedding belonging to the $K$ possible predicate classes. During training, $f_p$ minimizes a cross entropy objective to the true predicate class. Due to the inherent difficulties in modeling the object relationships using vision-only models, most successful approaches tend to be multi-modal~\citep{lu2016visual, zareian2020bridging, zellers2018neural}, i.e., exploit information in both the visual and language (semantic) domains. 

UVTransE \citep{hung2020contextual} utilizes the translational prior outlined above to obtain predictions using visual information, which is combined with language priors to get the final predicate estimate. While mathematically elegant, the performance has fallen short of expectations, particularly when compared to the success of translation-based methods in the knowledge graph embedding domain \citep{bordes2013translating}. One possible reason is that the translational properties may not be as applicable in the visual space when using only image embeddings for the subject, object, and union. Compounding these difficulties further, most common visual object relationship prediction benchmarks exhibit a long-tailed distribution \citep{krishna2017visual}, where some relationships are much more common than others. This imbalance further hinders our ability to effectively learn and comprehend less frequent but potentially meaningful entity interactions.

\begin{figure}[t]
    \centering
    \includegraphics[width = 0.90 \textwidth]{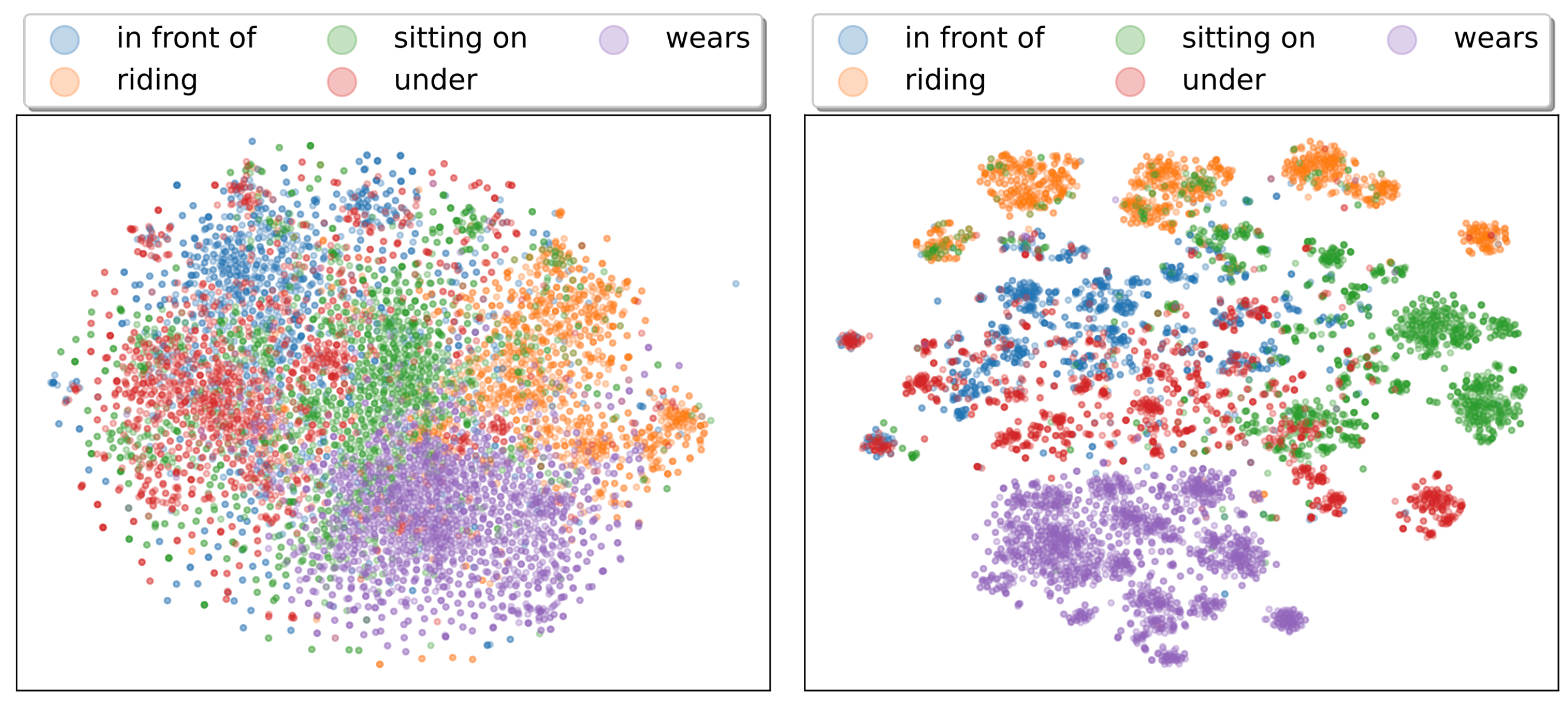}
    \caption{T-SNE visualization of the predicate representations from UVTransE trained with: (left) CLIP-based image embeddings for subject, object and union box regions; (right) CLIP-based image embedding for the union box, along with CLIP-based text embeddings for subject and object boxes.}
    \label{fig:tsne}
\end{figure}

To address these shortcomings of UVTransE contemporary research explores more sophisticated techniques to extract relationships, primarily focusing on the following approaches: \textit{1) Hybrid Modality Approach}: By combining language and visual information through complex model architectures, this approach harnesses the strengths of both, enabling robust relationship estimation; \textit{2) Graph-Based Methods}: Which utilizes trees, graphs, and hierarchical structures to model the global interaction of different entities in a scene to get better predicate representations; and
 \textit{3) Calibration/Bias Correction Techniques}: These techniques address model bias towards training data, improving the model's generalizability even to long-tailed predicates
 
 We contend that the complexities of current approaches are required to overcome the limitations of image embeddings and the challenges of using image-based embeddings for language-based relation estimation. We hypothesize that by leveraging carefully crafted embeddings from powerful VLMs like CLIP, the process of visual relationship estimation can be significantly simplified.  In the following sections, we will explore our methodology for obtaining these embeddings using CLIP.

\section{Approach}

\begin{figure}
    \centering
    \includegraphics[width = 0.75 \textwidth]{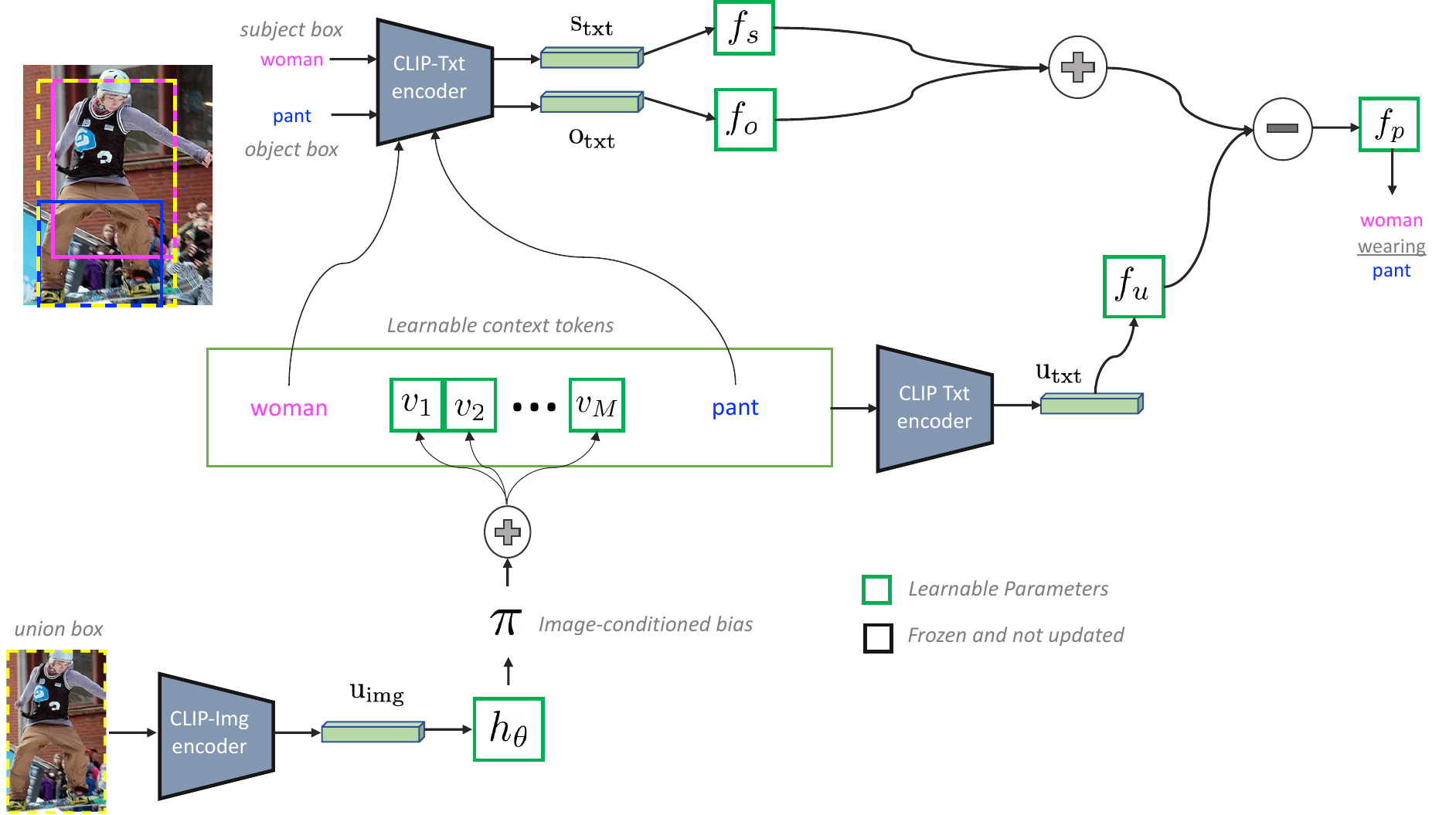}
    \caption{\textbf{CREPE}. An illustration of the proposed approach. CREPE uses learnable context vectors along with image-conditioned bias correction to obtain visually grounded text descriptors for an union image. Note that, the CLIP backbone is used to both perform the optimization for text prompt generation as well as producing the $(\mathrm{s}_{\footnotesize{\text{txt}}}, \mathrm{o}_{\footnotesize{\text{txt}}}, \mathrm{u}_{\footnotesize{\text{img}}})$ embeddings for UVTransE.}
    \label{fig:bd}
\end{figure}

In the task of object relation prediction, visual features and language priors are utilized to estimate the relationship between different objects within an image. Our main objective is to leverage the inherent visual and language priors of CLIP to obtain embeddings that simplify the estimation process and enhance its accuracy. Specifically, we focus on the UVTransE framework\citep{hung2020contextual} for relation prediction, mainly due to its simplicity, but our contributions are general enough to work with any framework. In this section, we introduce CREPE (CLIP Representation Enhanced Predicate Estimation), which emphasizes the use of text-based descriptors for the subject, object, and union bounding box regions. By incorporating CLIP embeddings for these text descriptors, CREPE aims to improve the reliability and precision of the relation prediction. For a visual depiction of CREPE, please refer to Figure \ref{fig:bd}.

\noindent \textbf{Representing Subject/Object Bounding Boxes.} A na\"ive way to leverage CLIP is to use the pre-trained image encoder to project the subject, object, and union bounding box regions into the joint embedding space, and use these representations to invoke UVTransE. We find that using CLIP merely as a pre-trained visual feature extractor does not provide any significant benefits over representations obtained from a conventional object detector, e.g., Faster-RCNN. We attribute this behavior to the fact that the translational prior produces weak predicate representations when both subject and object information are subtracted from the union image features.

In order to circumvent his challenge, we advocate for the use of coarse, text-based descriptors for both subject and object boxes that only convey the overall semantic concept. To this end, we leverage CLIP's pretrained text encoder to obtain subject and object embeddings from their respective object labels. As expected, when compared to the image representations, our modified embeddings are not specific to the characteristics of a given scene. When incorporated into the UVTransE framework, this leads to richer predicate embeddings, thereby resulting in significant performance gains compared to the na\"ive approach discussed earlier. Figure \ref{fig:tsne} shows the T-SNE visualization of the learned predicate representations from UVTransE invoked using image-based and object label-based representations for subject and object bounding boxes (using the appropriate encoder from CLIP). Note, in both cases, we used CLIP's image encoder to obtain embeddings for the union bounding box.

\noindent \textbf{Representing the Union Bounding Box.} While it is straightforward to obtain text representations for the subject and object regions, owing to the readily available labels (human-specified in our experiments or through an object detector in practice), creating such descriptors for the union image is challenging. This leads to our central question: can we automatically infer a text prompt for the union image that is visually grounded and also well-suited for UVTransE training?

At the outset, one straightforward approach is to manually engineer prompts~\citep{liu2022design} and optionally leverage Large-Language-Models~\citep{pratt2022does} to further refine them. However, it is challenging to construct meaningful prompts for all predicate classes in a dataset, and more important that the prompts offer sufficient visual grounding. Consequently, we opt for learnable prompting, wherein we automatically construct a prompt for the union image of the form \spo{subject}{learable-text-tokens}{object}. Here, the learnable text tokens are conditioned on the image, thus ensuring that the prompts are visually grounded, and the optimization is carried out by leveraging the inductive biases from CLIP. 

The learnable text tokens $[\mathrm{v}_1, \cdots , \mathrm{v}_M]$ are randomly initialized in the CLIP's token embedding space, i.e., the space where words or phrases from text sequences are converted into token embeddings. Bounded by start-of-sentence [SOS] and end-of-sentence [EOS] tokens, with a maximum length of 77 tokens, these embeddings are fed into a Transformer model to generate the final text representations. We allow the $M$ token vectors to be updated during training. Using these tokens, we construct the text prompt for an union image as $\{ \mathrm{s}_t-\{\mathrm{v}_i\}_{i=1}^M-\mathrm{o}_t\}$, where $\mathrm{s}_t$ and $\mathrm{o}_t$ are the subject and object token embeddings. While these context tokens are the same for all cases, we adopt an approach similar to the recently proposed CoCoOP~\citep{zhou2022conditional} to incorporate image-conditioned biases to learn image-specific prompts. We construct a non-linear MLP $h_\theta(.)$ which takes the CLIP image embedding for an union bounding box as input and outputs the bias $\mathrm{\pi} = h_\theta(\mathrm{u}_{\footnotesize{\text{img}}})$. In other words, each of the $M$ learnable text tokens are refined with the image-conditioned bias as $\mathrm{v}_m = \mathrm{v}_m + \mathrm{\pi}$. We refer to the prompt generation process using the notation $g_{\phi}(.)$, where the learnable parameters $\phi$ correspond to the context tokens $\{\mathrm{v}_m\}$ and parameters $\theta$ of the MLP $h_{\theta}$.

Although the idea of automatically inferring text descriptors for the union image is interesting, visually grounding these representations is difficult, since we lack ground truth text descriptors to guide the learning of the context tokens. To address this issue, we propose a novel contrastive training approach that utilizes the inductive biases from CLIP. We adopt a simple cross-modal retrieval strategy to generate pseudo labels for the union image and utilize them to construct negative samples in the contrastive training objective. More specifically, we first build an exhaustive vocabulary of all possible \spo{subject}{predicate}{object} triplets in the visual genome (VG) benchmark, and for each union image, select the most similar entry in the triplet vocabulary for each union image based on cosine similarity of CLIP embeddings.

Let us denote the CLIP image embedding of the union bounding box as $\mathrm{u}_{\footnotesize{\text{img}}}$, and the CLIP embedding for the learned text prompt for the union image as $\mathrm{u}_{\footnotesize{\text{txt}}} = \text{CLIP}_\text{txt}\big( g_\phi(\mathrm{s}_{\footnotesize{\text{txt}}}, \mathrm{o}_{\footnotesize{\text{txt}}}, \mathrm{u}_{\footnotesize{\text{img}}}) \big)$. We adopt the following contrastive objective to infer the parameters $\phi$:

\begin{equation}
\mathcal{L}_{LTD} = - \log \frac{\exp(\text{sim}(\mathrm{u}_{\footnotesize{\text{img}}}, \mathrm{u}_{\footnotesize{\text{txt}}}))}{\exp(\text{sim}(\mathrm{u}_{\footnotesize{\text{img}}}, \mathrm{u}_{\footnotesize{\text{txt}}})) + \exp(\text{sim}(\mathrm{u}_{\footnotesize{\text{img}}}, \mathrm{\hat{u}}_{\footnotesize{\text{txt}}}))}.
\end{equation}Here, $\text{sim}$ denotes the cosine similarity and $\mathrm{\hat{u}}_{\footnotesize{\text{txt}}}$ is the CLIP-text representation for the pseudo labels obtained via cross-modal retrieval with the VG triplet vocabulary. Note that, our training of the prompt generator is disconnected from UVTransE training. Intuitively, this optimization encourages text prompts that can more appropriately describe the content of the union image compared to its pseudo-label. Upon the completion of this contrastive training process $g_\phi(.)$, we incorporate the CLIP-text embeddings for subject, object and union boxes into UVTransE for predicate estimation.

\noindent \textbf{Post-hoc Calibration.} Similar to all existing methods in object relation estimation, we need to handle long-tailed predicate distribution challenge encountered in standardized benchmarks such as the Visual Genome (VG). To this end, we use a straightforward frequency-based correction technique, commonly referred to as Train-Est~\citep{elkan2008learning}. This method obtains an estimate of the label frequencies based on the (biased) prediction probabilities for each class from the trained UVTransE model. Subsequently, these frequency estimates are used to adjust the predicted probabilities for test samples during inference, with the hope of mitigating the skewness in the class label distribution. 

Let $P_{biased}(\mathrm{p}_k | (\mathrm{s}_{\footnotesize{\text{txt}}}, \mathrm{o}_{\footnotesize{\text{txt}}}, \mathrm{u}_{\footnotesize{\text{txt}}}))$ denote the biased probabilities for each predicate class $k$ from the UVTransE model. The frequency of each class label, denoted as $\beta_k$, is then computed as the average of the biased probabilities of that predicate over the entire dataset:
$$\beta_k = \frac{1}{|\mathcal{D}_i|} \sum_{(\mathrm{s}_{\footnotesize{\text{txt}}}, \mathrm{o}_{\footnotesize{\text{txt}}}, \mathrm{u}_{\footnotesize{\text{txt}}}) \in \mathcal{D}_i}^{|\mathcal{D}_i|} P_{biased}(\mathrm{p}_k | (\mathrm{s}_{\footnotesize{\text{txt}}}, \mathrm{o}_{\footnotesize{\text{txt}}}, \mathrm{u}_{\footnotesize{\text{txt}}})).$$Here, the term $\mathcal{D}_i$ is the set of all examples from the $k^{\text{th}}$ predicate class in the validation dataset and $|.|$ denotes the cardinality. The adjusted probabilities $P_{adj}(p_k | (\mathrm{s}_{\footnotesize{\text{txt}}}, \mathrm{o}_{\footnotesize{\text{txt}}}, \mathrm{u}_{\footnotesize{\text{txt}}}))$ can be then calculated as:

$$P_{adj}(\mathrm{p}_k | (\mathrm{s}_{\footnotesize{\text{txt}}}, \mathrm{o}_{\footnotesize{\text{txt}}}, \mathrm{u}_{\footnotesize{\text{txt}}})) = \frac{P_{biased}(\mathrm{p}_k | (\mathrm{s}_{\footnotesize{\text{txt}}}, \mathrm{o}_{\footnotesize{\text{txt}}}, \mathrm{u}_{\footnotesize{\text{txt}}}))}{\beta_k}.$$Through this simple protocol, Train-Est assists in reducing the bias introduced due to the long-tail distribution of the VG dataset, enhancing the model's performance across all classes. Interestingly, most state-of-the-art approaches have found that this na\"ive correction mechanism does not work effectively in practice and hence they opt for more sophisticated calibration strategies. However, we find that, even this simple procedure leads to high fidelity predicate estimates in CREPE.

\section{Experiments}

\subsection{Experiment Setup}

\textbf{Dataset:} Our approach is evaluated on the widely adopted Visual Genome (VG) dataset, a large-scale benchmark comprising 108,077 images with annotations for 75,000 object categories and 40,000 predicate categories. Notably, within these categories, there were 150 unique object categories and 50 unique predicate categories. Following common practice, we utilize a split of 57,772 training images, 26,440 test images, and 5,000 images for validation to assess the performance of our approach.

\textbf{Evaluation Protocol:} We evaluate our approach on the task of predicting the relations between legitimate pairs of entities based on ground truth bounding boxes and labels. To measure the performance, we utilize the standard Mean Recall@K (mR@K) metric. This metric is well suited for relation prediction tasks as it alleviates the issue with imbalanced sampling between different predicate categories. Unlike the vanilla recall metric, which can be influenced by the over-representation of certain predicates, mR@K aggregates the recall scores estimated independently for each individual predicate. Consequently, this has been found to provide a comprehensive evaluation of the model's ability to accurately classify predicates. In existing works, when reporting mR@K, it is typical to set K to be 50 or 100. However, with the advancement in the methodologies for predicate estimation, it is desirable to improve the performance at lower K values (better trade-off with precision). By reporting mR@K for smaller values, we gain insights into the model's ability to prioritize and retrieve the most relevant predicates, which is preferable in real-world use cases. Therefore, we strongly advocate for reporting mR@K for smaller values of K. In this work, we report the mean recall at $K$ set to $[5, 10, 15, 20, 50]$. 



\textbf{Implementation Details:} 
When implementing CREPE, $h_\theta(.)$ is designed as a simple $2-$layer MLP (Multi-Layer Perceptron), FCN -> ReLu -> FCN. The projection functions within UVTransE, namely $f_s$, $f_o$, and $f_u$, also use the same $2-$layer MLP deisgn. We train UVTransE using an SGD (Stochastic Gradient Descent) optimizer for 100 epochs. The initial learning rate is set at $1\mathrm{e}{-3}$, and we employ a non-monotone scheduler with learning rates alternating between $1\mathrm{e}{-3}$ and $1\mathrm{e}{-4}$ for predetermined epoch ranges. As for the training of $g_\phi(.)$, we use an SGD optimizer with a fixed learning rate of $2\mathrm{e}{-3}$, and the training extends for 500 epochs. Following conventional practice, we train our models with a separate \textit{no-relation} predicate class, but while computing the recall metrics we do not include this. For the vanilla UVTransE baseline, we employed a frozen Faster-RCNN~\citep{ren2015faster} to obtain the visual embeddings, in line with previous research~\citep{tang2019learning, tang2020unbiased, zellers2018neural, chen2019knowledge}. Note, this detector is equipped with the ResNeXt-101 FPN backbone and has been pre-trained~\citep{tang2020sggcode}. Our codes can be accessed at \hyperlink{https://github.com/LLNL/CREPE}{https://github.com/LLNL/CREPE}

\begin{figure}[h]
    \centering
    \includegraphics[width = 0.85 \textwidth]{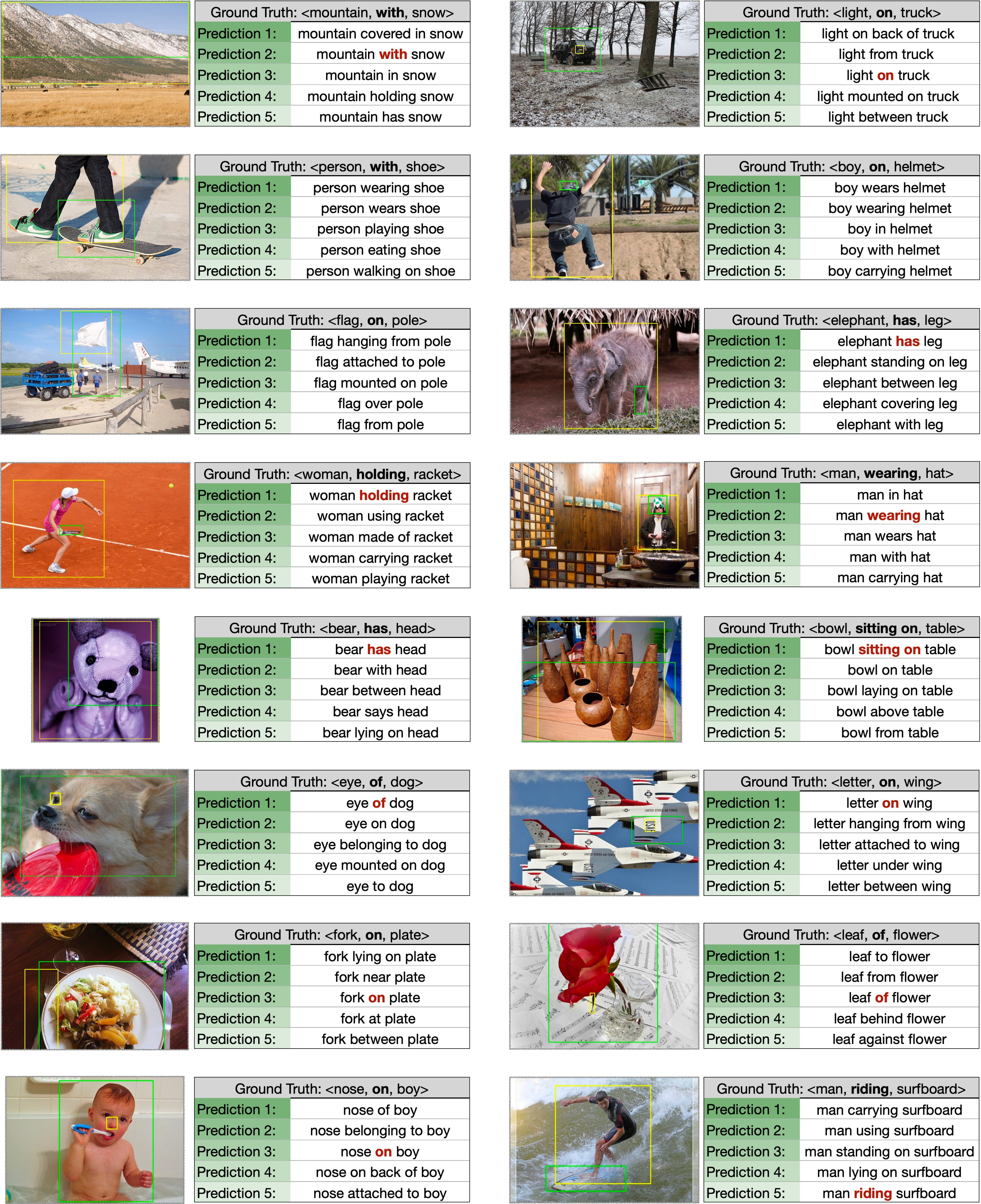}
    \caption{\textbf{Qualitative Results of CREPE}. Each sub-figure illustrates the relationship between the subject (yellow box) and the object (green box), accompanied by the top five predictions made by CREPE. The accurate prediction is emphasized in red. Notably, in the first column of the third row, although the ground truth label is \spo{flag}{on}{pole}, CREPE makes a more suitable prediction with \spo{flag}{hanging~from}{pole}, thus indicating that the evaluation metrics can be conservative.}
    \label{fig:examples}
\end{figure}

\subsection{Results}
We present a comprehensive comparison of CREPE with state-of-the-art methods in Table~\ref{tab: main}. It is worth emphasizing that our work is the first to report mR@{5, 10, 15}, and therefore, the results of other methods at these values are not available. 
Here, FREQ~\citep{zellers2018neural} denotes a basic frequency-based method, while UVTransE~\citep{hung2020contextual} (visual only) depends purely on CLIP-based image embedding for the union, subject, and object boxes and UVTranE (visual \& language) uses CLIP-based image embedding for the union box, along with CLIP-based text embeddings for subject and object boxes. Contrastingly, techniques such as KERN~\citep{chen2019knowledge}, IMP~\citep{xu2017scene}, MOTIFS~\citep{zellers2018neural}, VCTree~\citep{tang2019learning}, GB-Net~\citep{zareian2020bridging}, and SG-Transformer~\citep{yu2020cogtree} utilize sophisticated structural priors for a scene, e.g., graphs, trees, to augment the performance of relation prediction. We note that applying UVTransE solely on visual representations falls short of matching the performance of other state-of-the-art approaches.

Our main finding in this paper is that \emph{our proposed framework CREPE when combined with a simple translational model delivers state-of-the-art results, and this holds even for smaller values of the recall threshold}. 
This supports our hypothesis that text-based representations derived from multimodal networks like CLIP yield powerful priors for relationship prediction tasks. CREPE validates this claim by presenting a novel approach to extract such rich representations for relation prediction. Thus the right combination of model design and prompt engineering can lead to breakthroughs in performance, without requiring intricate structural designs.

\begin{table*}[t]
\centering
\caption{Predicate Estimation Performance: This chart compares the performance of our proposed CREPE method with other state-of-the-art methods on the Visual Genome (VG) dataset, using mean Recall@K (mR@K). The best performing method is highlighted in red, while the second best is in blue. It's worth noting that we are, to our knowledge, the first to report mR@ {5,10,15}, and hence those scores for other methods are not presented.}
\resizebox{0.99 \textwidth}{!}{
\begin{tabular}{@{}l|ccccc@{}}
\toprule
\multicolumn{1}{c|}{\textbf{Model}} &  \textbf{mR@5} & \textbf{mR@10} & \multicolumn{1}{c|}{\textbf{mR@15}} & \textbf{mR@20} & \textbf{mR@50}  \\ \midrule
FREQ~\cite{zellers2018neural}                          &        -.-       &       -.-         &            -.-                         &        8.3      & 13.0           \\
UVTransE (visual only) ~\cite{hung2020contextual}            &        -.-         &       -.-           &              -.-                         & 8.26           & 11.41  \\
\phantom{x}+ visual + language            &        -.-         &       -.-           &              -.-                         & 14.33           & 19.50  \\
KERN  ~\cite{chen2019knowledge}                            &        -.-       &       -.-         &            -.-                         &        -.-        & 17.7            \\
GB-Net   ~\cite{zareian2020bridging}                          &         -.-        &                 -.-                     & -.-              & 19.3                 \\
\midrule
IMP~\cite{xu2017scene}                                            &       -.-           &                  -.-                   & -.-              & 9.8  & 10.5 \\
\phantom{x}+ EBML~\cite{suhail2021energy}                  &        -.-       &       -.-         &            -.-                         &     -.-              & 11.8 \\
\phantom{x}+ PPDL~\cite{li2022ppdl}                  &        -.-       &       -.-         &            -.-                         &     -.-              & 24.8 \\
\midrule
MOTIFS   ~\cite{zellers2018neural}                  &          -.-               &          -.-       &               -.-                       & 11.5           & 14.6    \\
\phantom{x}+ TDE~\cite{tang2020unbiased}                   &        -.-       &       -.-         &            -.-                         &     18.5              & 25.5 \\
\phantom{x}+ PCPL~\cite{yan2020pcpl}                  &        -.-       &       -.-         &            -.-                         &     -.-              & 24.3 \\
\phantom{x}+ CogTree~\cite{yu2020cogtree}               &        -.-       &       -.-         &            -.-                         &     20.8             & 26.4 \\
\phantom{x}+ DLFE~\cite{chiou2021recovering}                  &        -.-       &       -.-         &            -.-                         &     22.1             & 26.9 \\
\phantom{x}+ BPL-SA~\cite{guo2021general}                &        -.-       &       -.-         &            -.-                         &     24.8             & 29.7 \\
\phantom{x}+ NICE~\cite{li2022devil}                  &        -.-       &       -.-         &            -.-                         &     -.-              & 29.9 \\
\phantom{x}+ SCR~\cite{kang2023skew}                   &        -.-       &       -.-         &            -.-                         &     25.9              & 31.5 \\
\midrule
VCTree~\cite{tang2019learning}                             &     -.-          &         -.-        &             -.-                         & 11.7           & 14.9                \\
\phantom{x}+ TDE~\cite{tang2020unbiased}     &        -.-       &       -.-         &            -.-                         &     18.4             & 25.4 \\
\phantom{x}+ PCPL~\cite{yan2020pcpl}    &        -.-       &       -.-         &            -.-                         &     -.-              & 22.8 \\
\phantom{x}+ CogTree~\cite{yu2020cogtree}    &        -.-       &       -.-         &            -.-                         &     22.0             & 27.6 \\
\phantom{x}+ DLFE~\cite{chiou2021recovering}   &        -.-       &       -.-         &            -.-                         &     20.8             & 25.3 \\
\phantom{x}+ BPL-SA~\cite{guo2021general}    &        -.-       &       -.-         &            -.-                         &     26.2             & 30.6 \\
\phantom{x}+ PPDL~\cite{li2022ppdl}    &        -.-       &       -.-         &            -.-                         &     -.-              & \secondbest{33.3} \\
\phantom{x}+ NICE~\cite{li2022devil}   &        -.-       &       -.-         &            -.-                         &     -.-              & 30.7 \\
\phantom{x}+ SCR~\cite{kang2023skew}   &        -.-       &       -.-         &            -.-                         &     \secondbest{27.7}              & \firstbest{33.5} \\
\midrule
SG-Transformer   ~\cite{yu2020cogtree}        &        -.-        &           -.-      &                    -.-                  & 14.8        & 19.2               \\
\phantom{x}+ CogTree~\cite{yu2020cogtree}    &        -.-       &       -.-         &            -.-                         &     22.9            & 28.4\\
\phantom{x}+ SCR~\cite{kang2023skew}    &        -.-       &       -.-         &            -.-                         &     27.0            & 32.2\\
\midrule
PE-Net (P)~\cite{zheng2023prototype}         &       -.-        &        -.-        &                 -.-                    & -.-              & 23.1                  \\
PE-Net~\cite{zheng2023prototype}               &      -.-         &            -.-    &                    -.-                 & -.-              & 31.5                  \\
\midrule
CREPE (ours)            &  \firstbest{27.79}         & \firstbest{31.12}          &   \firstbest{31.78}                               & \firstbest{31.95}          & 32.09       \\
 
\end{tabular}
}
\label{tab: main}
\end{table*}

\textbf{Key Findings:} CREPE achieves an mR@20 score of $31.95$, which is an improvement of 4.25 percentage points (or $15.3\%$ relative gain) compared to the second-best performing method, VCTree with SCR bias correction. More impressively, our framework's mR@5 performance at 27.79 
surpasses even the previously best mR@20 performance (via VCTree) as shown in Table~\ref{tab: main}. 
Thus there is no degradation in quality at much smaller recall thresholds, underscoring the effectiveness of our approach.

\begin{figure}[h]
    \centering
    \includegraphics[width = 0.80 \textwidth]{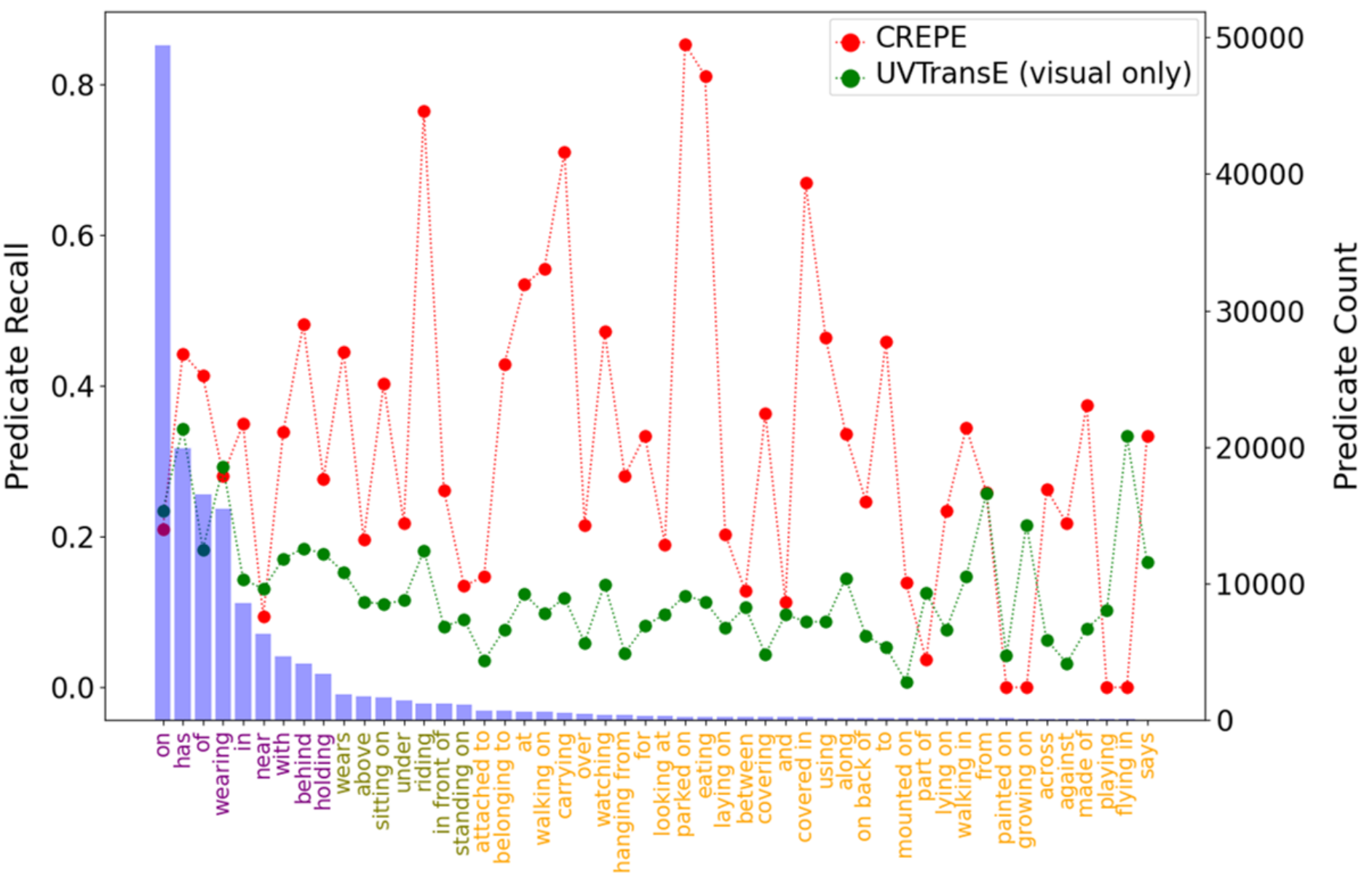}
    \caption{\textbf{Predicate level R@50}. The R@50 performance of two models CREPE, UVTransE (vision only), while also showing the frequency of each predicate. Predicates are color-coded based on their categories: `Head' (purple), `Mid' (olive), and `Tail' (orange). The recall values are shown as dotted lines, while the predicate frequencies are displayed as blue bars. }
    \label{fig:freq_plot}
\end{figure}

Figure~\ref{fig:freq_plot} shows that CREPE exhibits superior performance even in the tail predicates, when compared to UVTransE which uses vision-only representations. This highlights CREPE's ability to utilize the priors effectively learned from CLIP to low sample regimes.
Figure~\ref{fig:examples} demonstrates the quality of CREPE's predictions via some illustrative examples. 
For instance, considering the first image in the first column; Where the ground truth is \spo{mountain}{with}{snow} CREPE's top prediction is \spo{mountain}{covered~in}{snow} which is also appropriate; although CREPE matches the ground truth in the second best prediction. The image in the second example (1st row, 2nd column) shows that CREPE's top prediction wrongly judges the light to be at the back of the truck and a more benign prediction (in the top 5) matches the ground truth. Other examples similarly demonstrate CREPE's ability to handle diverse types of images and relations.

\section{Related work}

Due to the lack of a common latent space of representations, earlier methods relied on complex network architectures involving RNNs and graph neural networks to combine the image features and text labels. IMP~\citep{xu2017scene} proposed a graph convolutional neural network to capture higher-order dependencies and contextual information in the scene. MOTIFS~\citep{zellers2018neural} proposed the concept of neural motifs; they also addressed the challenge of fine-grained predicate classification and focused on frequently occurring patterns of relationships, which allowed for a better understanding of complex scenes and improving the accuracy of predicate classification. Graph RCNN~\citep{yang2018graph} integrates object detection and relationship classification by proposing a graph convolutional neural network architecture to model the dependencies between objects and relationships. VCTree~\citep{tang2019learning} used dynamic tree structures to capture the local and global visual contexts.

While significant progress has been made in this area, addressing the long tail problem, which refers to the scarcity of training examples for infrequent or rare predicates, remains a challenge. To that end, researchers have adopted an unbiased evaluation metric called Mean Recall, to focus on the long-tailed bias in SGG datasets, of which Visual Genome Dataset~\citep{krishna2017visual} is an exemplar. Gu et al.~\citep{gu2019scene} and Chen et al.~\citep{chen2019knowledge} integrated external knowledge into SGG models to address the bias of noisy annotations. Tang et al.~\citep{tang2020unbiased} proposed to adapt counterfactual causal inference to eliminate the prediction bias caused by the long-tailed data. Li et al.~\citep{li2022ppdl} propose a modified loss function called PPDL to weaken the model’s suppression for the tail classes. Suhail et al.~\citep{suhail2021energy} proposed an energy-based training method to alleviate the long-tailed bias. On the other hand, the CogTree~\citep{yu2020cogtree} model uses a hierarchical cognitive tree of predicates to locate the bias and enable focus on a small portion of easily confused predicates. Other works that proposed interesting ways to circumvent the prediction bias include Yang et al.~\citep{chen2019scene, zhang2019graphical, zareian2020weakly, yang2021probabilistic}. In contrast to existing approaches, we eliminate the need to explicitly learn the relation between image features and text descriptors through the use of VLMs, and more importantly achieve equally effective generalization even for smaller recall thresholds with the simple UVTransE model.

\section{Conclusion and Broader Impact}

In summary, our study unveils the untapped capabilities of vision-language models, particularly CLIP, in visual relationship prediction. Our model CREPE leverages text-based representations and a unique contrastive training strategy to achieve state-of-the-art performance in the field, while simultaneously tackling the long-tail issue prevalent in predicate occurrence distribution. The implications of our work are broad, with potential advancements in numerous applications like autonomous navigation and intelligent surveillance. However, the risk of misuse in scenarios such as invasive surveillance or biased decision-making underlines the necessity for cautious and ethical deployment of such technologies.

\acks{This work was performed under the auspices of the U.S. Department of Energy by the Lawrence Livermore National Laboratory under Contract No. DE-AC52-07NA27344. Supported by the LDRD Program under project 21-ERD-012. LLNL-CONF-851181. }


\bibliography{sample}

\newpage

\appendix

\section{Implementation Details}
\subsection{UVTransE Design}
Following~\citep{hung2020contextual}, we incorporate the location information obtained from the bounding boxes to augment the predicate features. We extract location features by encoding each entity's bounding box (representing a subject or object) into a 5-dimensional vector. For example, for the subject entity, we define $\mathbf{l}_s = \left(\frac{{x_s}}{{w_s}}, \frac{{y_s}}{{h_s}}, \frac{{x_s + w_s}}{{W_I}}, \frac{{y_s + h_s}}{{H_I}}, \frac{{A_s}}{{A_I}}\right)$, includes the center coordinates $(x_s, y_s)$ of the subject, width and height $(w_s, h_s)$, and the area of the region ($A_s$) within the image ($I$). Additionally, $W_I$ and $H_I$ represent the width and height of the image $I$. We can similarly define the location features for the object box, $\mathbf{l}_o$.

To represent union boxes, we compute a 9-dimensional feature vector, denoted as $\\
\mathbf{l}_{s\cup o} = \left(\frac{{x_s - x_o}}{{w_o}}, \frac{{y_s - y_o}}{{h_o}}, \log\left(\frac{{w_s}}{{w_o}}\right), \log\left(\frac{{h_s}}{{h_o}}\right), \frac{{x_o - x_s}}{{w_s}}, \frac{{y_o - y_s}}{{h_s}}, \log\left(\frac{{w_o}}{{w_s}}\right), \log\left(\frac{{h_o}}{{h_s}}\right), \frac{{A_u}}{{A_I}}\right)$. 

where $(x_s, y_s, w_s, h_s)$ and $(x_o, y_o, w_o, h_o)$ are the subject and object box coordinates, $A_u$ is the area of the union box. All the the location features ($\mathbf{l}_s$, $\mathbf{l}_o$, $\mathbf{l}_{s\cup o}$) are first concatenated into a 19-dimensional vector. This vector is then passed through a two-layer MLP  with an intermediate layer dimension of 32 and an output dimension of 16.

\subsection{Mean Recall@K Metric}
To compute the Mean Recall@K (mR@K), we adopted the implementation from the Scene-Graph-Benchmark.pytorch \footnote{https://github.com/KaihuaTang/Scene-Graph-Benchmark.pytorch} repository. The mR@K metric ensures that each predicate contributes equally to the total recall, irrespective of its frequency in the dataset, thus providing a balanced view of the performance of the model across all types of relations.

\subsection{Learning Rate Selection}
For UVTransE training, we use a non-monotone learning rate schedule for 100 epochs where the learning rate alternates between $1\mathrm{e}{-3}$ and $1\mathrm{e}{-4}$, and where the switching happens at the end of Epoch 15, Epoch 30, Epoch 45, and Epoch 80, and then stays at $1\mathrm{e}{-4}$ until the end of Epoch 100. For $g_\phi(.)$, we use a fixed learning rate of $2\mathrm{e}{-3}$ for the entire 500 epochs.

\section{Ablations}
To assess the importance of our learnable context tokens in generating a text representation from the union image, we conducted an ablation study directly using the pseudo labels from Cross-Modal Retrieval (CMR). Note, CMR retrieves the pseudo labels from an exhaustive vocabulary comprising all possible \spo{subject}{predicate}{object} triplets present in the Visual Genome (VG) benchmark. 

Our investigation considers two scenarios:
\begin{enumerate}
    \item For each union image, we selected the most similar entry from the triplet vocabulary, determined by the cosine similarity of the CLIP embeddings.
    \item It is possible that meaningful retrieval with a limited vocabulary for complex scenes can be challenging. Hence, instead of selecting a single entry, we selected the top K similar entries from the triplet vocabulary for each union image. The expectation is that the expanded search can capture additional contextual information that might have not been captured in Case 1. Since each union image is described using a set of text (triplet) descriptors, we incorporate an additional attention module into UVTransE that appropriately learns to weight each of the representations. Note that, this implementation is similar to SoTA attention-based, multiple-instance-learning formulations~\citep{ilse2018attention}.
\end{enumerate}The results of these two cases, alongside CREPE with learnable text tokens and UVTransE baselines, are illustrated in Table \ref{tab:ablation}. From the results, it is evident that the introduction of pseudo labels significantly enhances performance compared to the CLIP-based, visual-only and visual+language UVTransE baselines. However, we notice a consistent improvement in the predicate estimation performance when we allow additional context information, \textit{i.e.}, $K > 1$ and provides a $2.5\%$ gain in the mR@50 metric when compared to the case of $K = 1$. However, despite the improvement observed with pseudo labels, the performance of using learnable text tokens markedly outshines them. This outcome highlights the effectiveness of learnable text tokens, demonstrating its superiority over using a restricted vocabulary to identify a suitable triplet for the union image.

\begin{figure}[!h]
    \centering
    \includegraphics[width = 0.85 \textwidth]{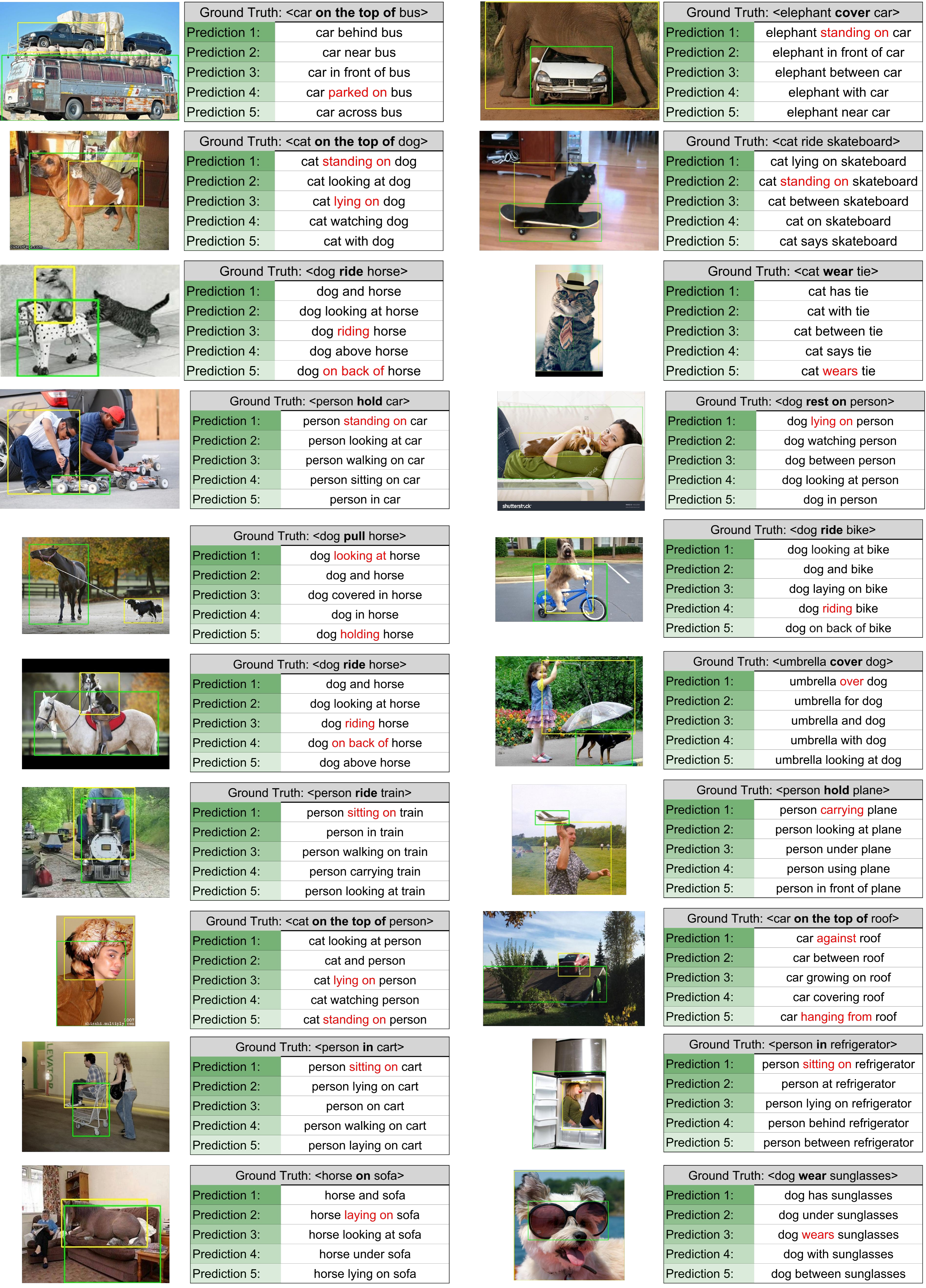}
    \caption{\textbf{Using CREPE to estimate predicates for the Unrel dataset.} Each sub-figure illustrates the relationship between the subject (yellow box) and the object (green box), accompanied by the top five predictions made by CREPE. The most suitable prediction is emphasized in red.}
    \label{fig:unrel}
\end{figure}

\begin{table}[h]
\centering
\caption{\textbf{Ablation On Union Representation.} This table compares the performance of UVTransE and CREPE with union image representations obtained using a variant of CREPE, that does not include learnable text tokens but directly uses the pseudo labels from Cross-Modal Retrieval for UVTransE. The best performing method is highlighted in red, while the second best is in blue.}
\begin{tabular}{@{}l|ccccc@{}}
\toprule
\multicolumn{1}{c|}{\textbf{Model}} &  \textbf{mR@5} & \textbf{mR@10} & \multicolumn{1}{c|}{\textbf{mR@15}} & \textbf{mR@20} & \textbf{mR@50}  \\ \midrule
UVTransE (visual only) ~\cite{hung2020contextual}            &       3.93         &       6.39           &              7.76                       & 8.26           & 11.41  \\
\phantom{x}+ language            &        6.33        &       9.85           &              12.30                         & 14.33           & 19.50  \\
\midrule
CREPE (w/o learnable prompting) & & & & & \\
\phantom{x}+ Pseudo Labels ($K = 1$)
&       8.49        &       12.45           &              15.15                         & 17.13           & 22.53  \\
\phantom{x}+ Pseudo Labels ($K = 3$)            &        9.85        &      14.40         &              \secondbest{17.36}                         & \secondbest{19.48}           & 24.61  \\
\phantom{x}+ Pseudo Labels ($K = 4$)            &        9.88         &       \secondbest{14.42  }         &              17.25                         & 19.12           & \secondbest{24.86}  \\
\phantom{x}+ Pseudo Labels ($K = 5$)            &        \secondbest{9.97}         &       13.84          &              17.04                         & 19.22           & 24.60  \\

\midrule
CREPE (ours)            &  \firstbest{27.79}         & \firstbest{31.12}          &   \firstbest{31.78}                               & \firstbest{31.95}          & 32.09       \\
 
\end{tabular}
\label{tab:ablation}
\end{table}

\section{CREPE performance on Unrel dataset}
In order to gauge the robustness and versatility of CREPE, we conducted an evaluation using the Unrel dataset~\citep{peyre2017weakly}. Note, we directly utilized CREPE trained on the Visual Genome (VG) dataset for this experiment. The Unrel dataset poses a unique challenge as it contains atypical relations that are not expected to occur in the VG benchmark, e.g., \spo{car}{on the top of}{bus} triplet. The performance on this benchmark will reveal the inherent biases in CREPE and how well it generalizes to novel object interactions. Since some of the predicates that the Unrel ground-truth contains are not present in the VG benchmark, we (qualitatively) consider the predicate estimate to be meaningful if the predicate returned by CREPE is relevant. We reserve a more rigorous quantitative analysis to future work.

The results for a small number of examples are showed in Figure ~\ref{fig:unrel}. We can clearly notice CREPE's ability to estimate these unconventional relations, despite not having been trained on such examples. For instance, consider the example depicted in the first row, second column, \spo{elephant}{cover}{car}. Here, CREPE's top prediction is \spo{elephant}{standing on}{car} which is an appropriate label for the image. Moreover, in the last row, second column, the triplet is \spo{dog}{wear}{sunglasses}. This scenario is doubly challenging: not only is the relationship atypical, but also the object \textit{sunglasses} is not present in the VG dataset. Despite these challenges, CREPE successfully identifies the correct relationship, demonstrating its robustness and generalizability.

\end{document}